\pdfoutput=1

\documentclass[11pt]{article}

\usepackage[final]{acl}

\usepackage{times}
\usepackage{latexsym}

\usepackage[T1]{fontenc}

\usepackage[utf8]{inputenc}

\usepackage{color}
\usepackage{tabularx}
\usepackage{algorithm}
\usepackage{algpseudocode}
\usepackage{tabu}
\usepackage{booktabs}
\usepackage{multirow}
\usepackage{subcaption} 
\usepackage{mathtools}

\usepackage{graphicx} 
\usepackage{float}
\usepackage{subfloat} 
\usepackage{amssymb}

\usepackage{hyperref}
\usepackage{tablefootnote}
\usepackage{xspace}

\usepackage{float}
\usepackage{amsmath}
\usepackage{bm}
\usepackage{algorithmicx}
\usepackage{algorithm}
\usepackage{algpseudocode}

\usepackage{arydshln}

\usepackage{amssymb}
\usepackage{pifont}
\usepackage{enumitem}

\usepackage{microtype}

\newcommand{\ours}{MCTS-RAG\xspace}

\newcommand{\ie}{\hbox{\emph{i.e.,}}\xspace}

\newcommand{\github}{\raisebox{-1.5pt}{\includegraphics[height=1.05em]{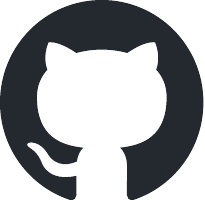}}\xspace}

\definecolor{YaleBlue}{RGB}{16, 42, 86}  
\definecolor{NYUPurple}{RGB}{134, 1, 175}  
\title{
\ours: Enhancing Retrieval-Augmented Generation with \\Monte Carlo Tree Search

}

\author{
Yunhai Hu$^{\hspace{.1em}{\textcolor{NYUPurple}{\boldsymbol{N}}}}$\thanks{~Equal contributions. Correspondence: Yilun Zhao (\texttt{yilun.zhao@yale.edu})} \quad
Yilun Zhao$^{\hspace{.1em}\textcolor{YaleBlue}{\boldsymbol{Y}}*}$ \quad
Chen Zhao$^{\hspace{.1em}\textcolor{NYUPurple}{\boldsymbol{N}}}$ \quad
Arman Cohan$^{\hspace{.1em}\textcolor{YaleBlue}{\boldsymbol{Y}}}$ \vspace{5pt}\\
$^{\textcolor{YaleBlue}{\boldsymbol{Y}}}$Yale University \quad 
$^{\textcolor{NYUPurple}{\boldsymbol{N}}}$New York University 
\vspace{5pt}
\\
\github ~~~\url{https://github.com/yale-nlp/MCTS-RAG}
}

\begin{document}
\maketitle

\begin{abstract}
We introduce \ours, a novel approach that enhances the reasoning capabilities of small language models on knowledge-intensive tasks by leveraging retrieval-augmented generation (RAG) to provide relevant context and Monte Carlo Tree Search (MCTS) to refine reasoning paths.
\ours dynamically integrates retrieval and reasoning through an iterative decision-making process. Unlike standard RAG methods, which typically retrieve information independently from reasoning and thus integrate knowledge suboptimally, or conventional MCTS reasoning, which depends solely on internal model knowledge without external facts, \ours combines structured reasoning with adaptive retrieval. This integrated approach enhances decision-making, reduces hallucinations, and ensures improved factual accuracy and response consistency.
The experimental results on multiple reasoning and knowledge-intensive datasets datasets (\ie ComplexWebQA, GPQA, and FoolMeTwice)
show that our method enables small-scale LMs to achieve performance comparable to frontier LLMs like GPT-4o by effectively scaling inference-time compute, setting a new standard for reasoning in small-scale models.
\end{abstract}

\section{Introduction}

\begin{figure}[t] 
    \centering
    \includegraphics[width=\linewidth]{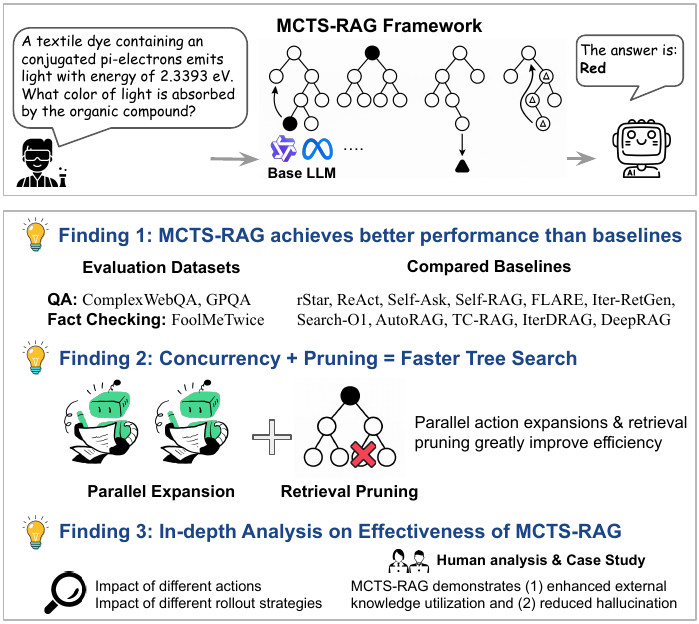} 
    \caption{
    Overview of the research. The top panel illustrates the proposed \ours framework, while the bottom panel summarizes three key findings from our experiments and analysis.
    }
    \label{fig:archi}
\end{figure}
Recent advancements in MCTS-based reasoning have demonstrated remarkable improvements in structured decision-making and logical inference~\cite{kocsis2006bandit, browne2012survey, xie2024monte}. The rStar framework~\cite{qi2024mutualreasoningmakessmaller}, for instance, has shown that systematic search and exploration can significantly enhance reasoning performance, enabling small-scale LMs (\ie models with up to 7B parameters)
to compete with much larger models. 
However, a key limitation of these approaches is their heavy reliance on internal knowledge, which hinders their effectiveness in knowledge-intensive tasks.

On the other hand, RAG has been widely used to solve knowledge-intensive tasks~\cite{lewis2020retrieval, karpukhin-etal-2020-dense, izacard-grave-2021-leveraging, zhao2025mmvu}, but its effectiveness with small-scale LMs remains limited.  
small-scale LMs struggle with query formulation and retrieved content comprehension, often generating vague queries and misinterpreting key details~\cite{fan2025miniragextremelysimpleretrievalaugmented}. 
Moreover, existing RAG systems do not dynamically adjust their retrieval strategies based on changing informational or reasoning requirements, which results in unnecessary or repetitive retrieval steps~\cite{li2024smartragselectionusingdeterminantal, gao2024retrievalaugmentedgenerationlargelanguage}. 
For example, when answering a multi-hop question like ``Which novel inspired the movie that won Best Picture in 1994?'', a standard retrieval system might retrieve documents about Forrest Gump (\ie Best Picture winner in 1994), but 
fail to recognize the need for additional reasoning or retrieval steps to establish the connection between Forrest Gump and the novel written by Winston Groom. This limitation arises because small-scale LMs often lack the ability to refine queries iteratively and integrate retrieved information into a coherent reasoning process. 


To address the aforementioned limitations, we propose MCTS-RAG, a novel framework that integrates MCTS’s reasoning and search capabilities with adaptive retrieval mechanisms. 
At a high level, \ours operates by iteratively refining both retrieval and reasoning through a search-based process. Given a query, it explores multiple reasoning paths, dynamically incorporating retrieval actions at key decision points. Retrieved knowledge is then used to evaluate intermediate states, and beneficial retrieval pathways are reinforced through backpropagation. This structured search mechanism ensures that the model efficiently acquires and utilizes relevant information for more accurate reasoning.
To further enhance efficiency, \ours employs parallel expansion and retrieval pruning strategies during the search, reducing redundant computation while maintaining search quality.
In contrast to prior approaches, by integrating retrieval with search-based reasoning, \ours is able to systematically explore relevant knowledge and reason over it to obtain the correct answer.


\ours has the following key features: 
\textbf{Improved reasoning accuracy:} New retrieval actions enable SLMs to acquire external knowledge and enhance the quality of question answering (\S\ref{sec:action-space}). 
\textbf{Optimized query formulation:} 
The refinement process ensures that each query focuses on specific information needs, improving the effectiveness of retrieval query generation (\S\ref{retrieve-process}). 
\textbf{Enhanced retrieval quality:} Reflecting on and summarizing retrieved information helps reduce semantic discrepancies and ensures alignment with the core problem (\S\ref{retrieve-process}).
\textbf{High efficiency:} Parallel expansion and retrieval pruning reduce redundant computation during search, greatly improving inference efficiency without compromising performance (\S\ref{sec:efficiency}).

\ours demonstrates superior performance on various knowledge-intensive benchmarks, 
including ComplexWebQA (CMQA)~\cite{talmor2018web}, GPQA~\cite{rein2024gpqa}, and FoolMeTwice (FMT)~\cite{eisenschlos-etal-2021-fool}. Specifically, it achieves over 20\% improvement with Llama 3.1-8B and ~6\% with Qwen2.5-7B on CWQA, roughly 15\% and 10\% gains on GPQA, and over 10\% (Llama) and 4\% (Qwen) on FMT, while outperforming other competitive baselines.
Our efficiency analysis demonstrates that \ours delivers the best overall trade-off compared to other baseline systems, achieving the highest accuracy with moderate latency and token cost.


\begin{figure*}[!t] 
    \centering
    \includegraphics[width=0.99\textwidth]{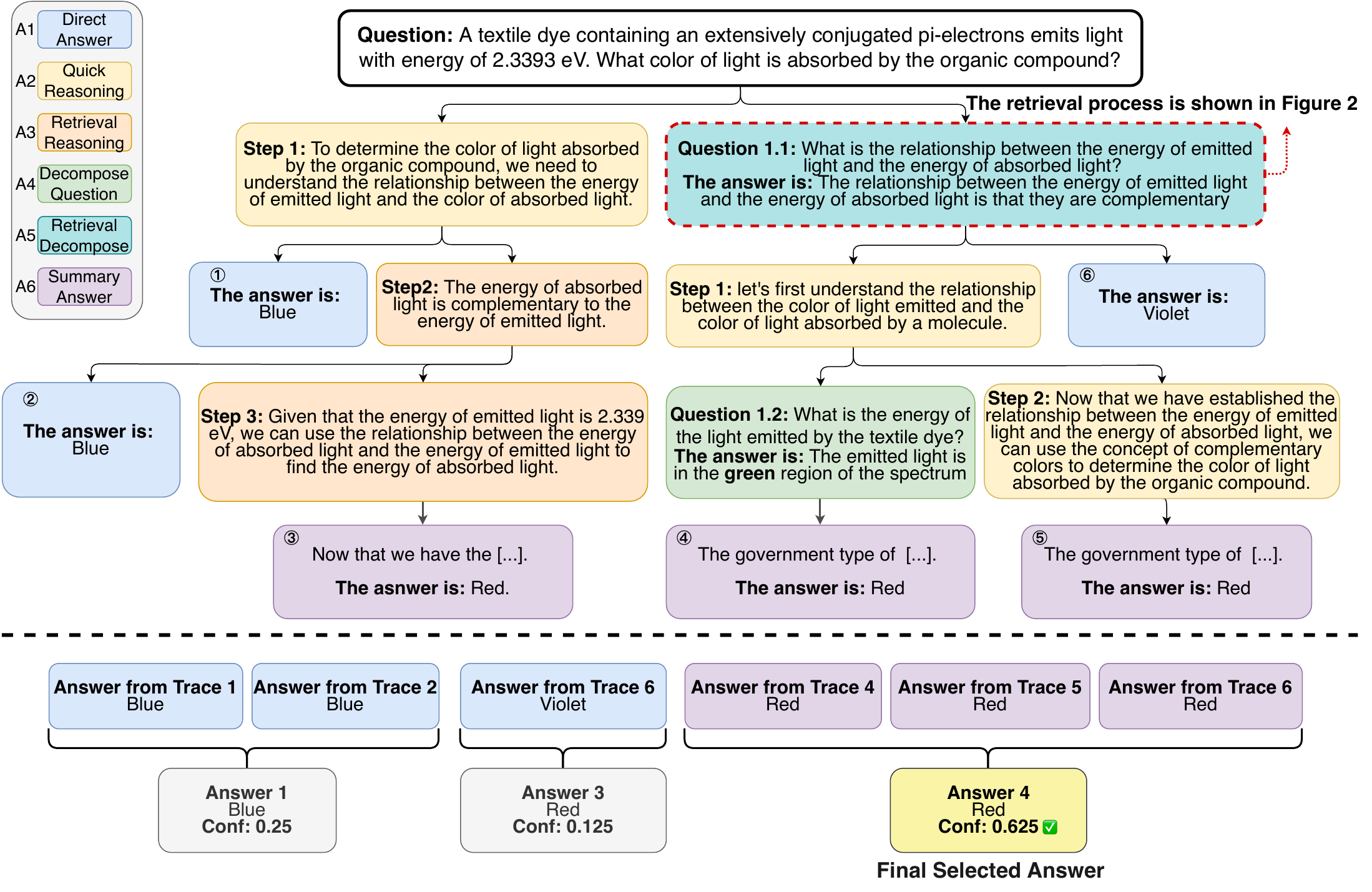} 
    \caption{
    An illustration of MCTS-RAG workflow for answering the question sampled from ComplexWebQA. 
    }
    \label{fig:reasoning-example}
\end{figure*}

\section{Related Work}
\paragraph{Inference-time Scaling.}
Inference-time scaling enhances reasoning without modifying model parameters by optimizing computational allocation during generation. A core approach involves reasoning diversification and selection: generating multiple candidates \cite{wang2023selfconsistency} and choosing optimal outputs via voting \cite{liang-etal-2024-encouraging} or verifier-guided ranking \cite{cobbe2021training}. Structured search algorithms, such as beam search \cite{xie2024self} and tree-of-thought frameworks \cite{yao2023tree}, explicitly model reasoning paths. Recently, Monte Carlo Tree Search (MCTS) has been applied to balance exploration and exploitation in reasoning tasks, iteratively refining solutions through selection, expansion, simulation, and backpropagation \cite{hao-etal-2023-reasoning}. Further, integrating MCTS with LLMs using value functions \cite{zhang2024rest} or predefined reasoning heuristics \cite{qi2024mutualreasoningmakessmaller} has improved efficiency in mathematical reasoning and code generation.

\paragraph{Retrieval-Augmented Generation.} 
The RAG system enhances LLMs in knowledge-intensive tasks by incorporating external information. Query optimization techniques, including expansion and transformation, improve retrieval quality \cite{ma2023query,jagerman2023query}. Iterative retrieval methods, such as IRCoT \cite{trivedi-etal-2023-interleaving} and ITER-RETGEN \cite{shao2023enhancing}, refine retrieval and generation. 
LLM-driven retrieval strategies, such as WebGPT \cite{nakano2021webgpt} and Toolformer \cite{schick2023toolformer}, have demonstrated notable improvements in efficiency by leveraging large language models to interact with external tools or search engines, thus streamlining the process of gathering relevant data. Meanwhile, self-reflection mechanisms in systems like Self-RAG \cite{asaiself,islam2024open} and Auto-RAG \cite{yu2024auto} further enhance retrieval relevance by employing iterative introspection to refine intermediate outputs. 
More recently, ReARTeR~\cite{sun2025rearter} improves retrieval by reasoning over multiple answer candidates with mutual evaluation. However, its sequential processing and full candidate scoring lead to high latency.
RAG-Star~\cite{jiang2024rag} uses a fixed search tree for token-level retrieval control, but lacks dynamic pruning and does not support concurrent reasoning paths.
In contrast, MCTS-RAG supports parallel expansion of diverse reasoning strategies and incorporates lightweight retrieval pruning to skip unnecessary external calls. This leads to better efficiency without sacrificing answer quality.

\section{\ours}
\subsection{Preliminaries}
rStar~\cite{qi2024mutualreasoningmakessmaller} is a recently proposed self-consistency framework designed to enhance the reasoning capabilities of language models without requiring additional fine-tuning or reliance on stronger teacher models. rStar achieves this by breaking down the reasoning process into two distinct yet interconnected phases: \textit{generation} and \textit{discrimination}. In the \textbf{Generation Phase}, the model proactively explores multiple reasoning trajectories through human-like reasoning actions, including step-by-step inference and question decomposition. Subsequently, the \textbf{Discrimination Phase} evaluates these candidate reasoning paths, selecting and refining them to identify the most logically consistent and accurate responses.

However, the original rStar framework is limited by its inability to dynamically acquire external knowledge, restricting its performance in knowledge-intensive queries. In addition, it suffers from significant latency, often requiring 4–5$\times$ the inference time compared to Standard RAG methods.
To address the inherent limitations of rStar, we propose an integrated reasoning framework that combines the iterative reasoning capabilities of rStar with RAG. At a high level, our approach builds on the iterative generative-discriminative structure of rStar and introduces additional operations specifically designed to facilitate dynamic external knowledge retrieval. This enables the language model to seamlessly integrate relevant external information into its reasoning process, significantly improving factual accuracy and decision robustness.
The following subsections detail the proposed \ours framework.


\subsection{Action Space Definition} \label{sec:action-space}
We design a set of discrete actions at each MCTS decision point: A1–A3 from rStar~\cite{qi2024mutualreasoningmakessmaller}, along with two new RAG-related actions A4 and A5 and a summary action A6, enabling dynamic knowledge acquisition and enhanced reasoning synergy for improved decision-making.


\begin{enumerate}[label=\textbf{A\arabic*:}, leftmargin=*] 
\item \textbf{Direct Answer:} 
Provide an immediate response based on existing reasoning or previously known context, suitable for straightforward queries or when additional analysis is unnecessary.

\item \textbf{Quick Reasoning:} 
Execute rapid, incremental reasoning steps based on the current context, ideal for exploratory paths or preliminary judgments to efficiently guide the search.

\item \textbf{Decompose Question:} 
Break complex queries into smaller, manageable sub-questions, allowing for clearer problem-solving pathways and improved reasoning efficiency, particularly beneficial for multi-part or intricate problems.

\item \textbf{Retrieval Reasoning:} 
Actively retrieve relevant knowledge from internal or external sources before proceeding with the next reasoning step, critical for queries requiring supplementary information or when existing context is incomplete.

\item \textbf{Retrieval Decompose:} 
Integrate both decomposition and retrieval, first breaking down complex questions and then acquiring relevant knowledge to solve individual sub-problems. This action is highly effective for queries involving detailed context-dependent sub-questions.

\item \textbf{Summarized Answer:} 
Generate a concise, structured summary that synthesizes results from previous reasoning and retrieved information, providing coherent and comprehensive responses especially useful for queries that demand summarization or integration of multifaceted information.

\end{enumerate}



Algorithm~\ref{alg:reward-update} illustrates the procedure for updating each action’s reward. To further enhance exploration, we employ Upper Confidence Bound for Trees (UCT)~\cite{kocsis2006bandit} in our MCTS framework—a crucial method that balances exploitation and exploration. The UCT formula is:

\begin{small}
\[
\text{UCT}(s,a) = \bar{Q}(s, a) + C \cdot \sqrt{\frac{\ln N(s)}{N(s, a)}}
\]
\end{small}

\noindent where \(\bar{Q}(s, a) = \frac{Q(s, a)}{N(s, a)}\) is the average reward for action \(a\) in state \(s\), with \(Q(s, a)\) as the cumulative reward and \(N(s, a)\) as the visit count. \(N(s)\) is the total number of visits to state \(s\). \(C\) is the exploration constant, controlling the balance between exploitation and exploration.

Within \ours, search depth limits how many levels are expanded from the root node to control the search range, while the number of rollouts indicates how many times the simulation is run from a selected node until termination or a preset limit to estimate its value. By running simulations within a controlled depth and updating node statistics via UCT, MCTS effectively balances exploration and exploitation with finite computational resources, continuously refining its search strategy.

\begin{algorithm}[t]
\caption{$R(s, a)$ Computation and Update}
\label{alg:reward-update}
\small
\begin{algorithmic}[1]
\Require State $s$, action $a$, completions $\{o_1, \dots, o_K\}$, log-likelihoods $\{\ell_1, \dots, \ell_K\}$
\Ensure Representative answer $o^*$, confidence $\mathrm{Conf}(o^*)$, reward $R(s,a)$

\State Initialize empty clusters $\mathcal{C} \gets \emptyset$
\For{$j = 1$ to $K$}
    \State $o_j \gets$ $j$-th completion
    \State matched $\gets$ \textbf{False}
    \For{each representative $r$ in $\mathcal{C}$}
        \If{$\texttt{Equiv}(o_j, r)$}
            \State Add $o_j$ to cluster $\mathcal{C}_r$
            \State matched $\gets$ \textbf{True}
            \State \textbf{break}
        \EndIf
    \EndFor
    \If{not matched}
        \State Create new cluster $\mathcal{C}_{o_j} \gets \{o_j\}$
    \EndIf
\EndFor

\State Let $\mathcal{O}^* \gets$ majority cluster with size $n^*$
\State Select representative $o^* \in \mathcal{O}^*$
\State $\mathrm{Conf}(o^*) \gets \frac{n^*}{K}$
\State $R(s, a) \gets \frac{1}{n^*} \sum_{o_j \in \mathcal{O}^*} \ell_j$

\State $Q(s,a) \gets Q(s,a) + R(s,a)$
\State $N(s,a) \gets N(s,a) + 1$

\Return $o^*$, $\mathrm{Conf}(o^*)$, $R(s,a)$
\end{algorithmic}
\end{algorithm}

\subsection{Retrieval Process}
\label{retrieve-process}
Our approach dynamically retrieves information within an evolving MCTS reasoning environment, enabling timely and relevant integration of external knowledge. The model autonomously determines when retrieval is required, generates targeted queries, and critically integrates external knowledge to improve reasoning accuracy. By interweaving retrieval with reasoning, we streamline information flow and produce concise yet informative outputs. If previously retrieved data adequately answers the current reasoning step—determined by checking whether the information satisfies predefined accuracy thresholds or resolves open reasoning paths—the model foregoes additional retrieval, thus avoiding redundancy.
\begin{enumerate}[label=\textbf{R\arabic*:}, leftmargin=*]  
    \item \textbf{Query Generation:} If a knowledge gap is detected, the model generates search queries.
    \item \textbf{Query Execution:} External retrieval tools are used to obtain the most relevant information.
    \item \textbf{Knowledge Reflection:} Retrieved data is evaluated for relevance and consistency to determine its inclusion in the reasoning process.
    \item \textbf{Summary Answer:} Refined information is integrated, enabling the model to answer subquestions or advance reasoning.
\end{enumerate}
This interleaved retrieval process ensures that the model’s reasoning is continuously updated and validated against external data, thereby reducing errors and enhancing the robustness of final output.
\begin{figure}[t] 
    \centering
    \includegraphics[width=0.45\textwidth]{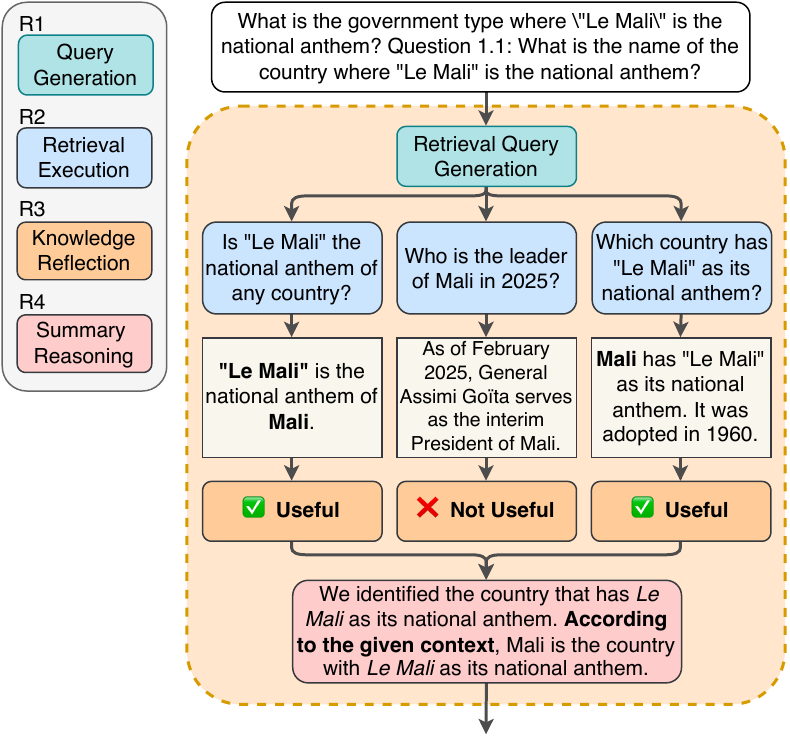} 
    \caption{
    An illustration of \ours retrieval process (\ie R1-R4) within one step of the retrieval decomposition action highlighted in \autoref{fig:example}.
    }
    \label{fig:example}
\end{figure}

\subsection{Determing Final Answer}
At the conclusion of the MCTS exploration (illustrated in the bottom part of
\autoref{fig:example}), the best answer is selected through a voting mechanism and consistency analysis over candidate solutions. Specifically, each reasoning trajectory obtained from the MCTS yields a candidate answer $c_j$, resulting in a candidate answer set $\mathcal{C} = \{c_1, c_2, \dots, c_M\}$. These candidate answers are grouped into a set of unique answers $\mathcal{A} = \{a_1, a_2, \dots, a_N\}$ based on semantic consistency. The final score for each unique answer $a_k$ is computed as the sum of the rewards of all candidates grouped under $a_k$, where the reward of each candidate $c_j$ is the product of rewards for all nodes along its corresponding reasoning trajectory.


\begin{equation}
\mathrm{Score}(a_k) =\frac{\sum_{c_j \in \mathcal{C}(a_k)} \mathrm{Reward}(c_j)}
    {\sum_{c_j \in \mathcal{C}} \mathrm{Reward}(c_j)}
\end{equation}

\noindent The best answer is then determined as 

\begin{equation}
a^*
= \arg\max_{\,a_k\,\in\,\mathcal{A}} \,\mathrm{Score}(a_k),
\end{equation}

\noindent ensuring that the most frequent and consistent reasoning trajectory is chosen. Essentially, our approach operates as a reward‐accumulation voting scheme: by normalizing and aggregating the trajectory rewards of semantically consistent candidate clusters, it guarantees that the selected answer is both the most coherent and the highest‐scoring.

\section{Experiment Setup}
\subsection{Evaluation Benchmark} 
We evaluate \ours and other competitive baseline systems on three complex reasoning tasks: (1) \textbf{ComplexWebQA (CWQA)}~\cite{talmor2018web}, which requires multi-step reasoning over web-based queries; (2) \textbf{GPQA}~\cite{rein2023gpqa}, which tests knowledge-intensive science question answering; and (3) \textbf{FoolMeTwice (FMT)}~\cite{eisenschlos2021fool}, a challenging fact-checking benchmark that assesses the model’s ability to verify factual claims.

\subsection{Baseline Systems}
Beyond \textbf{CoT prompting}, \textbf{rStar}, and \textbf{Standard RAG}, we also compare \ours with several recent RAG variants, including: \textbf{ReAct}~\cite{yao2023react} alternates between reasoning and retrieval, allowing the model to dynamically refine its understanding based on external evidence. \textbf{Self-Ask with Search (Self-Ask)}~\cite{press-etal-2023-measuring} with Search decomposes complex queries into subquestions, retrieves relevant external information, and synthesizes the answers to enhance multi-step reasoning. \textbf{Search-O1}~\cite{search-o1} executes a single retrieval step before generating an answer, limiting its ability to iteratively verify information. 
\textbf{Self-RAG}~\cite{asai2024selfrag} extends standard RAG by generating and issuing its own retrieval queries at each decoding step, enabling deeper multi-hop evidence gathering. 
\textbf{FLARE}~\cite{jiang2023active} employs an active learning strategy to select the most informative passages for retrieval, improving answer relevance with fewer retrieval calls. 
\textbf{Iter-RetGen}~\cite{shao2023enhancing} alternates retrieval and generation phases in multiple passes, refining responses through iterative editing. 
\textbf{AutoRAG}~\cite{kim2024autoragautomatedframeworkoptimization} automates retrieval pipeline configuration and model selection to optimize end-to-end retrieval-augmented generation performance. 
\textbf{TC-RAG}~\cite{jiang2024tcragturingcompleteragscasestudy} integrates Turing-complete control flows into RAG, supporting complex multi-step reasoning via case-based retrieval loops. 
\textbf{IterDRAG}~\cite{yueinference} introduces dynamic retrieval–generation loops that adapt retrieval strategies based on intermediate model outputs. 
\textbf{DeepRAG}~\cite{guan2025deepragthinkingretrievalstep} interleaves deep reasoning modules between retrieval steps to enhance multi-hop inference capabilities. \textbf{ReARTeR}~\cite{sun2025rearter} uses a reward model and explanations to improve step-by-step reasoning with MCTS. \textbf{RAG-Star}~\cite{jiang2024rag} combines MCTS planning with external retrieval to guide and verify multi-step reasoning. 
To ensure fair comparisons, we use the same base LLMs—Qwen2.5-7B and Llama 3.1-8B—for all the evaluated systems.
\subsection{Implementation Details}
\paragraph{RAG Setup.} 
To maintain consistency across methods, we use a shared retrieval corpus and identical retriever configurations.
We employ the Bing Search Engine and LangChain for retrieval, with Bing offering extensive and up-to-date web information and LangChain providing modular support for retrieval-augmented generation workflows.
For the CWQA dataset, we collect approximately 100K web snippets retrieved via Bing using question templates; these snippets are dynamically retrieved at inference time based on the input question. For GPQA, we construct a static corpus comprising 80K passages sampled from Wikipedia and 60K web documents retrieved from Bing, totaling 140K documents. The retriever encodes questions and selects top-ranked documents from this hybrid source. For FMT, a domain-specific dataset, we use its original associated documents (about 30K passages) provided as part of the benchmark and treat them as the retrieval pool. In all cases, top-10 retrieved documents are fed into the reasoning module for answer generation and verification.
This setup ensures that variations in performance stem from differences in reasoning mechanisms.

\paragraph{\ours Setup.} 
To facilitate structured reasoning, we configure our setup with a \emph{rollout} of 4, allowing multiple steps of reasoning expansion. Each query can be decomposed into at most two subquestions, ensuring a controlled breakdown of complex queries. We set the \emph{maximum reasoning depth} to 5, enabling deep but efficient multi-hop reasoning.
Moreover, in order to reduce latency during tree expansion, we implement concurrent action evaluations: different reasoning actions at a given search node are expanded in parallel. This design leverages asynchronous computation to significantly speed up MCTS traversal without sacrificing the fidelity of action-value estimation.

\begin{table}[!t]
\centering
\small
\setlength{\tabcolsep}{1.6pt}
\renewcommand{\arraystretch}{1.1}
\begin{tabular}{lcccccc}
\toprule
\multirow{2}{*}{\textbf{Methods}} & \multicolumn{3}{c}{\textbf{Qwen2.5-7B}} & \multicolumn{3}{c}{\textbf{Llama 3.1-8B}} \\
\cmidrule(l{2pt}r{5pt}){2-4} \cmidrule(l{5pt}r{2pt}){5-7} 
 & CWQA & GPQA & FMT & CWQA & GPQA & FMT \\
\midrule
CoT              & 34.6 & 35.0 & 57.3 & 27.7 & 28.7 & 56.5 \\
~~GPT-4o         & 54.4 & 53.0 & 55.4 & 54.5 & 53.0 & 55.4 \\ 
~~Qwen2.5-72B    & 44.5 & 40.6 & 58.4 & 44.6 & 40.6 & 58.4 \\
rStar            & 55.4 & 32.3 & 55.9 & 37.6 & 28.7 & 56.4 \\
\noalign{\vskip 0.2ex}\hdashline\noalign{\vskip 0.2ex}
Standard RAG     & 44.2 & 40.6 & 58.4 & 35.6 & 31.7 & 51.5 \\
~~GPT-4o         & 59.4 & 54.9 & 61.4 & 59.4 & 54.9 & 61.4 \\ 
~~Qwen2.5-72B    & 48.5 & 43.1 & 59.4 & 48.8 & 46.2 & 59.8 \\
ReAct            & 45.5 & 41.6 & 62.4 & 47.5 & 49.3 & 55.4 \\
Self-Ask         & 44.6 & 42.6 & 60.9 & 44.6 & 52.8 & 58.4 \\
Self-RAG         & 46.2 & 43.1 & 61.1 & 47.1 & 53.9 & 60.2 \\
FLARE            & 50.1 & 45.3 & 62.6 & 50.1 & 56.6 & 62.1 \\
ReARTeR          & 51.8 & 46.4 & 
63.3 & 51.4 & 57.1 & 64.3 \\
Iter-RetGen      & 52.2 & 47.5 & 63.1 & 52.3 & 57.6 & 63.2 \\
Search-O1        & 49.5 & 54.5 & 64.4 & 54.6 & 58.8 & 65.9 \\
AutoRAG          & 57.2 & 55.1 & 66.1 & 57.2 & 59.3 & 66.9 \\
TC-RAG           & 57.7 & 55.5 & 66.7 & 59.6 & 62.3 & 68.7 \\
RAG-Star         & 58.1 & 57.9 &
67.4 & 60.1 & 66.3 & 69.8 \\
IterDRAG         & 59.3 & 58.2 & 67.1 & 61.1 & 67.2 & 71.1 \\
DeepRAG          & 60.9 & 61.3 & 66.9 & 62.3 & 69.8 & 72.9 \\
\noalign{\vskip 0.2ex}\hline\noalign{\vskip 0.5ex}
\textbf{\ours}   & \textbf{61.4} & \textbf{64.6} & \textbf{68.3} & \textbf{67.3} & \textbf{71.3} & \textbf{73.8} \\
\bottomrule
\end{tabular}
\caption{Answer accuracy of MCTS-RAG and other methods (with and without retrieval modules).}
\label{tab:results}
\end{table}
\subsection{Main Findings}
\autoref{tab:results} compares reasoning methods on CWQA, GPQA, and FMT for Llama 3.1-8B and Qwen2.5-7B. Our approach consistently outperforms baselines, demonstrating strong multi-step reasoning and retrieval capabilities. On CWQA, it achieves over a 20\% gain with Llama 3.1-8B and around 6\% with Qwen2.5-7B. Similarly, it surpasses competitors on GPQA by roughly 42\% and 32\%, respectively, benefiting from refined verification strategies. On FMT, it leads by over 17\% with Llama 3.1-8B and 12\% with Qwen2.5-7B, proving its resilience against misleading distractors. These results highlight our method’s superior generalization and efficiency, especially in fact-checking and science-related tasks. 
Compared to baselines like Standard RAG, ReAct, Self-Ask, Search-O1, TC-RAG, IterDRAG, and DeepRAG, our structured multi-step reasoning can retrieve and process evidence more accurately, and on average we improve the performance by about 14\% over the baseline under three datasets. Compared to rStar, \ours enables broader retrieval, extracting critical insights while minimizing hallucinations, achieving an average improvement of 17\%. 

\subsection{Efficiency Analysis}
\label{sec:efficiency}

\begin{figure}[t] 
    \centering
    \includegraphics[width=0.48\textwidth]{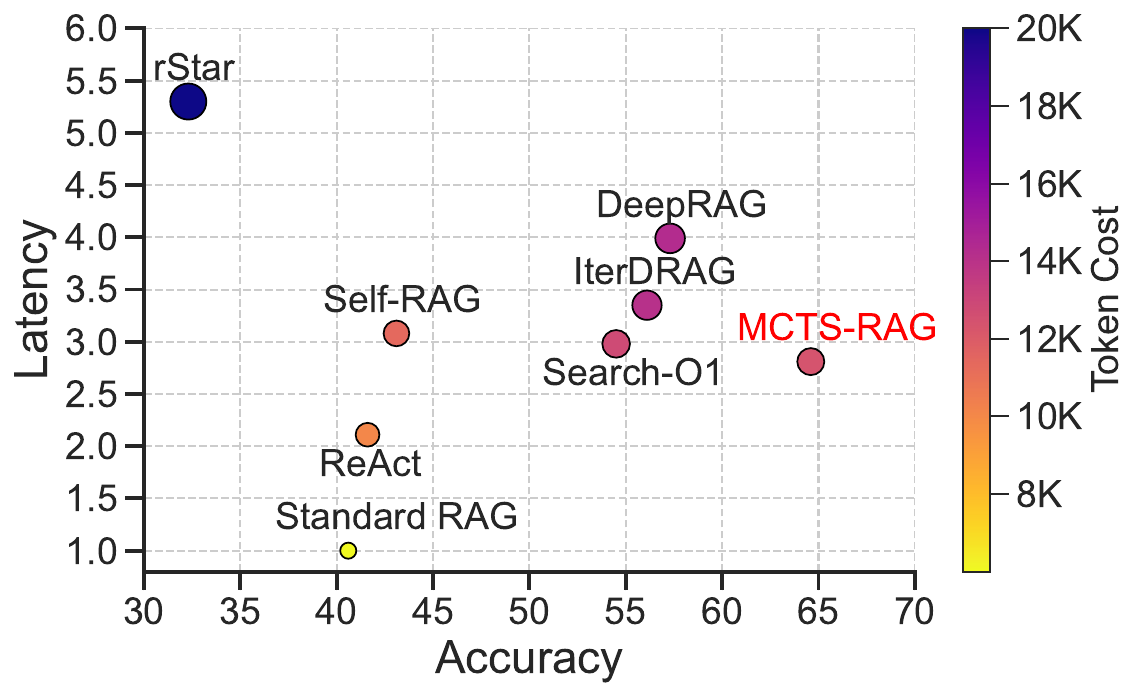} 
    \caption{
    Comparison of different methods on GPQA using Qwen2.5-7B. The horizontal axis shows accuracy, the vertical axis shows relative latency the decoding time relative to the Standard RAG baseline, and marker size and color correspond to average token generated.}
    \label{fig:method_comparison}
\end{figure}
As shown in \autoref{fig:method_comparison}, the standard RAG system serves as the baseline with normalized latency (1.0×) but also the lowest accuracy (40.6\%). ReAct improves accuracy to 42.6\% at the cost of 2.0× latency and slightly lower token consumption (8,884). Self-RAG further increases accuracy to 43.1\% with 2.8× latency and 11,000 tokens. Search-O1 balances accuracy (54.5\%) and latency (2.8×) with moderate token cost (11,300), making it suitable for latency-sensitive scenarios. IterDRAG (56.1\%, 3.3×, 12,500 tokens) and DeepRAG (57.3\%, 4.0×, 13,700 tokens) trade additional latency for marginal accuracy gains. rStar performs poorly (32.3\%, 5.5×, 19,000 tokens), indicating low efficiency. 
MCTS-RAG delivers the best overall trade-off, achieving the highest accuracy (64.6\%) with moderate latency (2.8×) and token cost (11,892).

\autoref{tab:retrieval} presents the token cost and relative latency for Qwen2.5-7B under various retrieval and rollout configurations. While MCTS-RAG consumes more tokens per query than the Standard RAG baseline (11,892 vs.\ 9,993 tokens in the “Enable All” vs.\ “Disable A4” settings), it achieves significantly better latency efficiency, running only 2.81$\times$ slower than RAG despite a 19\% increase in token cost. In contrast, rStar’s latency grows nearly linearly with rollout depth, since it cannot parallelize across multiple reasoning branches.

This improvement stems from two key optimizations in MCTS-RAG:  
(1) \textbf{Parallel Expansion of Reasoning Actions:}  during each rollout step, candidate subquestions (“actions”) are generated and verified in parallel, rather than sequentially as in rStar. This concurrency amortizes the overhead of verification across multiple branches, yielding higher token throughput per unit time.
(2) \textbf{Dynamic Pruning of Unnecessary Retrievals:}
Our framework allows the model to autonomously determine whether a retrieval step is needed. If the current context suffices, the model skips retrieval entirely. To improve both accuracy and efficiency, this pruning decision is guided by two lightweight yet effective signals: (i) a \emph{retrieval necessity signal}, where the model first decides, based on the current prompt and context, whether external retrieval is required, thus avoiding unnecessary queries; and (ii) a \emph{candidate similarity signa}, where multiple reasoning candidates are compared at each step and branches with abnormally low consistency are pruned, as they typically indicate hallucinations or reasoning failures. This implicit gating mechanism reduces redundant queries without introducing additional pruning modules. Together, these signals enhance reasoning stability, reduce computational cost, and are particularly valuable in resource-constrained environments. 

As shown in the rollout analysis (\autoref{tab:retrieval}), increasing the rollout number from 4 to 16 raises token cost from 11,892 to 28,972, but latency only increases from 2.8$\times$ to 4.5$\times$ relative to Standard RAG. This sublinear latency growth confirms that our concurrent action expansion and adaptive branch pruning jointly deliver a favorable trade-off: modest increases in token consumption yield diminishing increments in inference time, substantially outperforming both Standard RAG and rStar in overall efficiency.

\begin{table}[!t]
\centering
\small
\renewcommand{\arraystretch}{1.1}
\setlength{\tabcolsep}{2.5pt}
\begin{tabular}{lccccc}
\toprule
\textbf{Settings} & \textbf{CWQA} & \textbf{GPQA} & \textbf{FMT} & \textbf{Token} & \textbf{Latency} \\
\midrule
\multicolumn{6}{c}{\textbf{\textit{Analysis of Retrieval Modules}}} \\\noalign{\vskip 0.8ex}
Disable A1       & 61.3 & 64.6 & 68.1 & 11300 & 2.7$\times$ \\
Disable A2       & 60.5 & 63.5 & 67.5 & 15460 & 3.1$\times$ \\
Disable A3       & 60.7 & 63.9 & 67.6 & 16050 & 3.3$\times$ \\
Disable A4\&A5   & 55.5 & 32.3 & 50.4 &  8884 & 2.4$\times$ \\
Disable A4       & 55.7 & 36.3 & 55.9 &  9993 & 2.6$\times$ \\
Disable A5       & 56.2 & 44.1 & 62.4 &  9714 & 2.5$\times$ \\
\noalign{\vskip 0.2ex}\hdashline\noalign{\vskip 0.2ex}
Enable All       & 61.4 & 64.6 & 68.3 & 11892 & 2.8$\times$ \\
\midrule
\multicolumn{6}{c}{\textbf{\textit{Analysis of Rollout Numbers}}} \\\noalign{\vskip 0.8ex}
4 rollout        & 61.4 & 64.6 & 68.3 & 11892 & 2.8$\times$ \\
\noalign{\vskip 0.2ex}\hdashline\noalign{\vskip 0.2ex}
8 rollout        & 64.4 & 63.7 & 68.1 & 16963 & 3.2$\times$ \\
12 rollout       & 68.7 & 75.2 & 69.4 & 21860 & 3.9$\times$ \\
16 rollout       & 71.2 & 84.3 & 74.1 & 28972 & 4.5$\times$ \\
\bottomrule
\end{tabular}
\caption{Accuracy of Qwen2.5-7B under various retrieval and rollout settings. Token: avg. generated tokens; Latency: relative decoding time vs standard RAG.}
\label{tab:retrieval}
\end{table}

\subsection{Fine-grained Analysis}
We conduct an ablation by disabling different retrieval modules (A4, A5, or both) to gauge their impact on overall performance. In addition, we vary the number of rollouts from 4 to 16 to investigate how deeper search affects accuracy and efficiency.
\autoref{tab:retrieval} shows the results.

\paragraph{Impact of Different Actions.} 
Retrieval actions, especially A4 and A5, are key for multi-step reasoning. Enabling all retrievals boosts GPQA (+32.3\%) and FMT (+17.9\%). Disabling A5 improves GPQA (+11.8\%) and FMT (+12.0\%) over disabling A4, suggesting A4’s stronger role. CWQA sees minimal impact (+5.9\%). These findings highlight retrieval trade-offs and the importance of recursive evidence aggregation.
\paragraph{Impact of Different Rollout Strategies.} 
More rollouts enhance performance. Specifically, increasing from 4 to 8 slightly aids CWQA (+3.0\%), while 8 to 12 boosts GPQA (+11.5\%). Scaling to 16 further improves GPQA (+9.1\%) and FMT (+4.7\%), showing the value of iterative reasoning.



\subsection{Human Analysis and Case Study}
To better understand the strengths and limitations of \ours, we conduct a comprehensive analysis of its successful cases in comparison to baseline methods, along with a thorough error analysis.
\paragraph{Successful Case Analysis.} Our case study reveals the following two key improvements introduced by \ours:
(1) \textbf{Enhanced External Knowledge Utilization}:
Compared to other reasoning methods, \ours achieves higher accuracy, primarily due to its richer reasoning space and more effective utilization of external knowledge. \autoref{fig:draft_ee} clearly illustrates how Monte Carlo Tree Search tightly integrates reasoning and retrieval processes, significantly enhancing the quality and richness of information used during reasoning, thereby substantially improving inference accuracy.
(2) \textbf{Reduced Hallucination Risks}:
Moreover, \ours mitigates hallucination risks through detailed and explicit reasoning steps. On one hand, the explicit reasoning pathways enable the model to more accurately interpret retrieved external knowledge, reducing errors arising from ambiguity or misunderstanding (as illustrated in  \autoref{fig:correct_hallucination} in Appendix). On the other hand, these thorough reasoning procedures generate clearer and more contextually relevant queries, thus improving the precision of external information retrieval (as illustrated in \autoref{fig:reasoning_query} in Appendix). 
\paragraph{Error Case Analysis.}
Our human analysis identifies the following three primary error types in \ours:
(1) \textbf{Amplification Error}: As illustrated in \autoref{fig:mcts-error-example}, early retrieval errors in \ours can be magnified, causing incorrect information to dominate subsequent reasoning and ultimately leading to a incorrect final answer.
(2) \textbf{Factual Confusion}:
We reveal that semantic mismatches between retrieved text and the reasoning process can lead to conflations or hallucinations. \autoref{fig:factual-confusion-error-example} presents details on how semantically divergent retrieval results can lead to incorrect final answers. 
(3) \textbf{Information Overload}: Excessive additional information in MCTS-RAG can cause certain reasoning paths to deviate from the original question, leading to incorrect conclusions. \autoref{fig:information-overload-error-example} presents a detailed example of some reasoning paths that prioritize irrelevant aspects.

\section{Conclusion}
The work introduces \ours, which integrates MCTS with RAG to improve multi-step reasoning accuracy and reliability. 
\ours not only enables flexible formulation of high-quality retrieval queries but also refines the reasoning path through iterative tree exploration, thus reducing hallucinations caused by shallow retrieval or simplistic reasoning. 
To further enhance practicality, we introduce several efficiency optimizations, including parallel expansion and retrieval pruning, significantly reducing inference latency without compromising accuracy. 
Experimental results demonstrate that \ours achieves strong performance on complex reasoning, knowledge-enhanced scientific QA, and fact-checking tasks.


 \section*{Limitations and Future Work}
MCTS-RAG integrates MCTS-based reasoning and RAG to enhance reasoning capabilities, but several errors persist. Amplification errors occur when early retrieval mistakes propagate through search iterations. Factual confusion arises from semantic mismatches leading to incorrect reasoning. Information overload happens when excessive retrieval results cause reasoning to deviate from the target.
Additionally, search latency remains a challenge, as deep MCTS search trees significantly increase reasoning time, particularly with multiple retrieval steps. Action selection complexity arises because the optimal choice among A1-A6 depends on query difficulty, necessitating a more adaptive decision mechanism. While our current heuristic-based action space design provides a practical balance between task complexity and potential reward, it may not optimally adapt to all query types. We further acknowledge that dynamically adjusting the action space could improve flexibility in more complex environments, but may also introduce hallucinations or instability, particularly for small-capacity models. Inefficient expansion occurs when MCTS explores unnecessary branches due to a lack of effective pruning based on retrieval confidence or early error detection. Addressing these issues is essential for improving efficiency and reasoning accuracy.

We encourage future work to focus on optimizing search efficiency by developing adaptive action selection strategies, confidence-based retrieval filtering, and error-aware pruning mechanisms to improve MCTS exploration. Additionally, integrating reinforcement learning for dynamic search policy refinement may further enhance reasoning accuracy. Exploring ways to refine dynamic action-space adjustment while mitigating instability for small models is another promising direction. Addressing these challenges will contribute to the development of more robust and scalable reasoning models, bridging the gap between retrieval-based methods and human-like problem-solving.

\bibliography{anthology,custom}

\appendix

\clearpage
\onecolumn
\section{Prompts for Each Action}

\subsubsection*{A1 (Direct Response)}
\textbf{Template:}\\
\texttt{A chat between a curious user and an AI assistant. The assistant gives step-by-step solutions to the user's questions. In the end of the assistant's response, a final answer must be given in the format of "The answer is: <ANSWER>.", where <ANSWER> should be a concise answer.}\\[5pt]
\textbf{Usage Example:}\\
\texttt{\{examples\}}\\[5pt]
\textbf{Instruction:}\\
\texttt{\{instruction\}}\\[5pt]
\textbf{Note:}\\
Please answer in a complete sentence.

\hrulefill

\subsubsection*{A2 (One-Step Reasoning)}
\textbf{Template:}\\
\texttt{A chat between a curious user and an AI assistant. The assistant gives step-by-step solutions to the user's questions with each step numbered. At the final step, a conclusive answer must be given in the format "The answer is: <ANSWER>.", where <ANSWER> should be a concise answer.}\\[5pt]
\textbf{Instruction:}\\
\texttt{\{instruction\}}\\[5pt]
\textbf{Note:}\\
\texttt{Let's think step by step.}

\hrulefill

\subsubsection*{A3 (Decompose Answer)}
\textbf{Template:}\\
\texttt{Given a question, decompose it into sub-questions. For each sub-question, provide an answer in one complete sentence ending with "The answer is ". When the original question is answerable, start the sub-question with "Now we can answer the question: <original question>".}

\hrulefill

\subsubsection*{A4 (Transform Retrieve Query)}
\textbf{Template:}\\
\texttt{Given a question, generate a search query that would help gather information to answer it. Your goal is to formulate a query that retrieves useful evidence or additional details relevant to the question. The query should be specific enough to ensure that the search results are both relevant and helpful. Please answer in one complete sentence, starting with "The query is: <your retrieve query>".}\\[5pt]
\textbf{Question:}\\
\texttt{\{question\}}

\hrulefill

\subsubsection*{A5 (Reflect Retrieved Knowledge)}
\textbf{Template:}\\
\texttt{A chat between a curious user and an AI assistant. The assistant evaluates whether the retrieved information is relevant to the search query and sufficient to answer the question. Please provide a concise evaluation in one complete sentence, starting with "Evaluation:".}\\[5pt]
\textbf{Instruction:}\\
\texttt{Please assess if the retrieved information is related to the query and can be used to answer the question.}

\hrulefill

\subsubsection*{A6 (Summarize Answers)}
\textbf{Template:}\\
\texttt{Analyze the provided \textbf{Knowledge} and extract key information relevant to the \textbf{Original Question}. Present your analysis in a concise and organized format.}\\[5pt]
\textbf{Input:}\\
\texttt{- \textbf{Original Question:} \{original\_question\} \\
- \textbf{Knowledge:} \{retrieved\_context\}}\\[5pt]
\textbf{Output Format:}\\
\texttt{Key Points: Point 1: Relevant information; Point 2: Relevant information; Point 3: Relevant information...}\\[5pt]
\textbf{Requirement:}\\
\texttt{The output must be a single line summarizing all key points in one sentence without redundant description.}

\twocolumn

\clearpage
\onecolumn
\section{Error Analysis}

\begin{figure}[h] 
    \centering
    \includegraphics[width=0.49\textwidth]{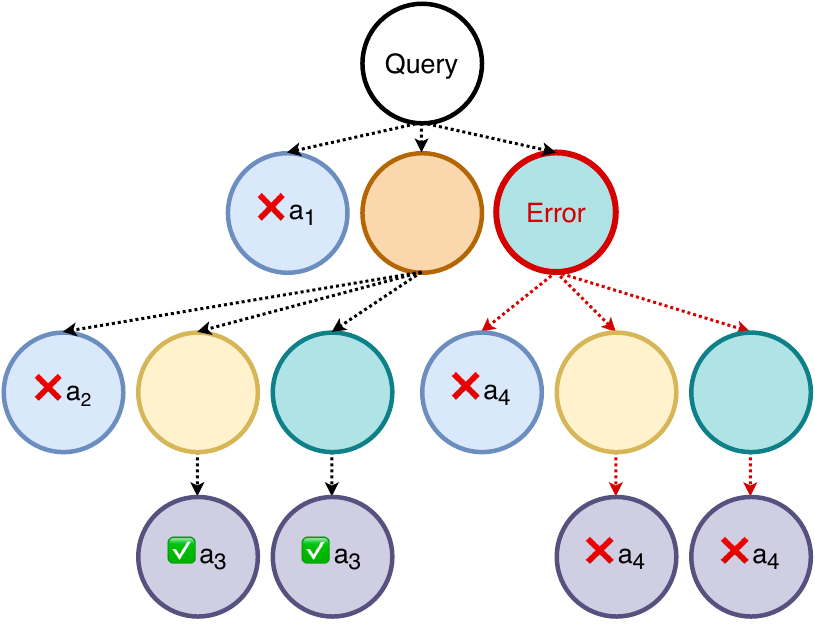} 
    \caption{
    An illustration of MCTS Amplification Error. Early MCTS retrieval errors amplify mistakes, leading to a final answer favoring incorrect paths.
    }
    \label{fig:mcts-error-example}
\end{figure}

\begin{figure}[h] 
    \centering
    \includegraphics[width=0.85\textwidth]{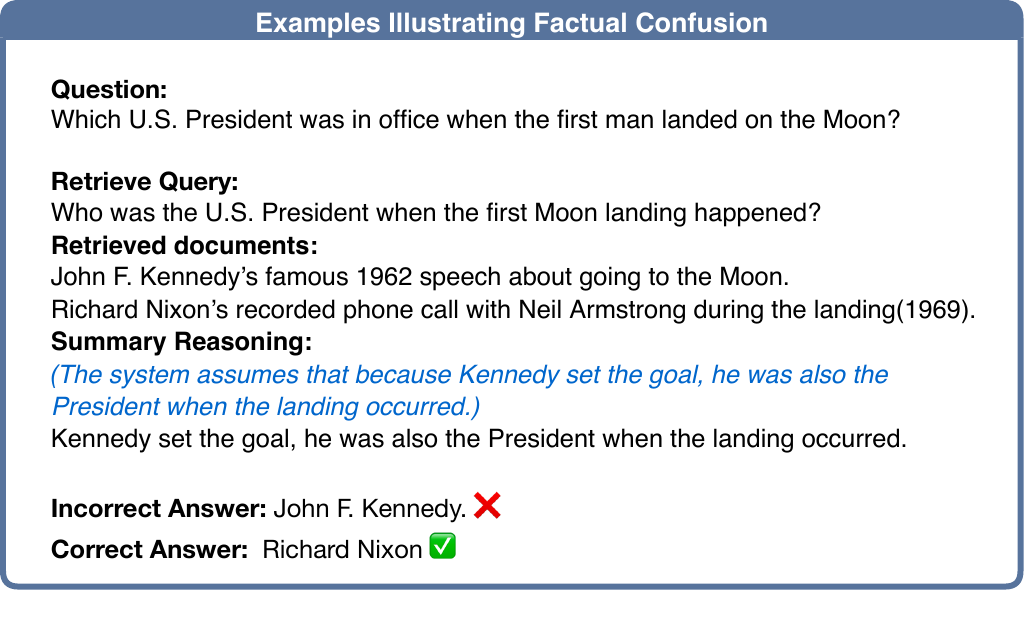} 
    \caption{
    An illustration of Factual Confusion. Wrong understanding of the relationship between project launch and moon landing, leading to wrong answers.
     }
    \label{fig:factual-confusion-error-example}
\end{figure}
\clearpage 
\begin{figure}[!h] 
    \centering
    \includegraphics[width=0.85\textwidth]{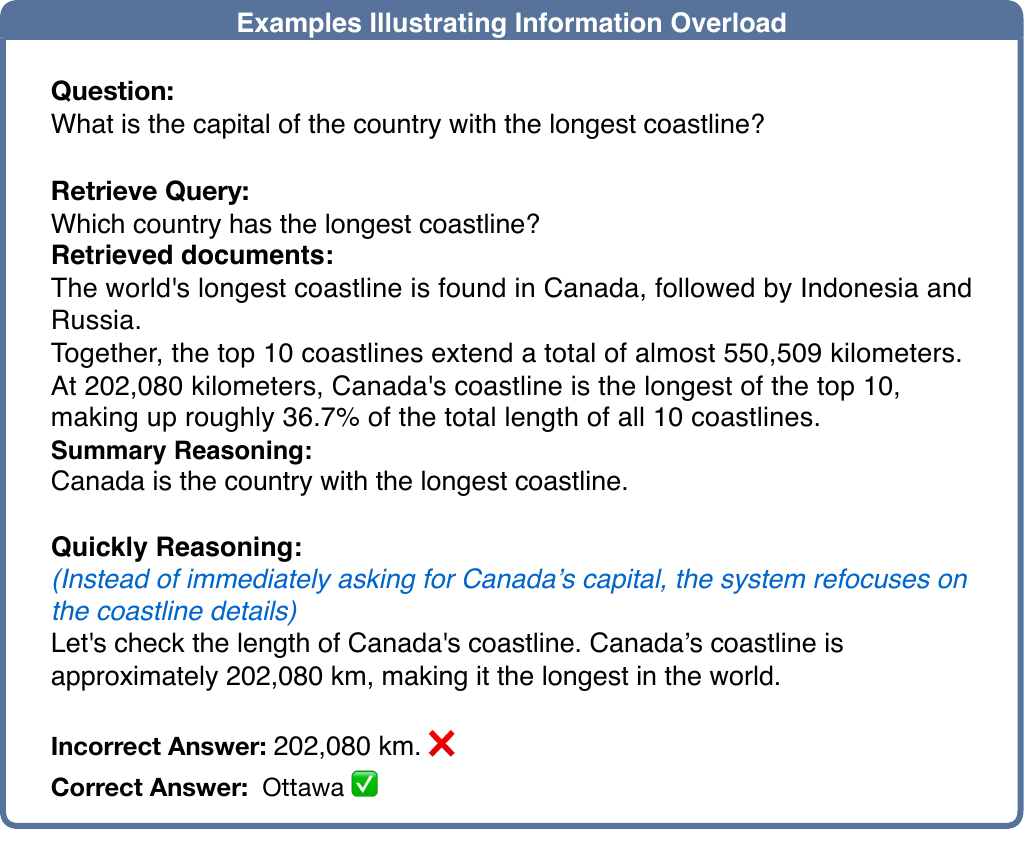} 
    \caption{
    An illustration of Information Overload. Too much coastline information, resulting in the model answering the coastline length instead of the capital city.
    }
    \label{fig:information-overload-error-example}
\end{figure}
\begin{figure}[!h] 
    \centering
    \includegraphics[width=0.99\textwidth]{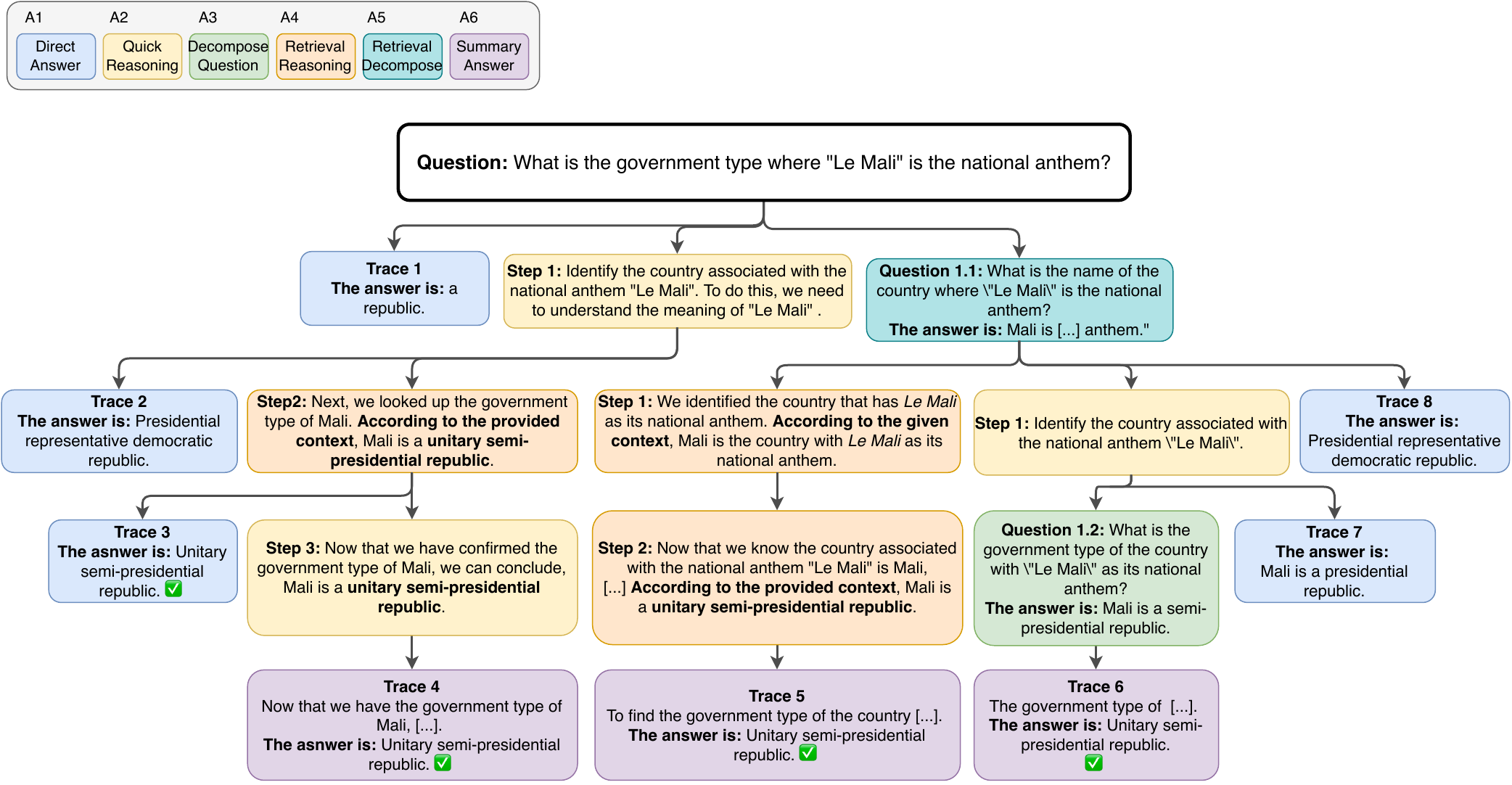} 
    \caption{
   Illustration of how \ours achieves a rich reasoning space and tightly integrates reasoning with retrieval.
    }
    \label{fig:draft_ee}
\end{figure}
\begin{figure}[!h] 
    \centering
    \includegraphics[width=0.85\textwidth]{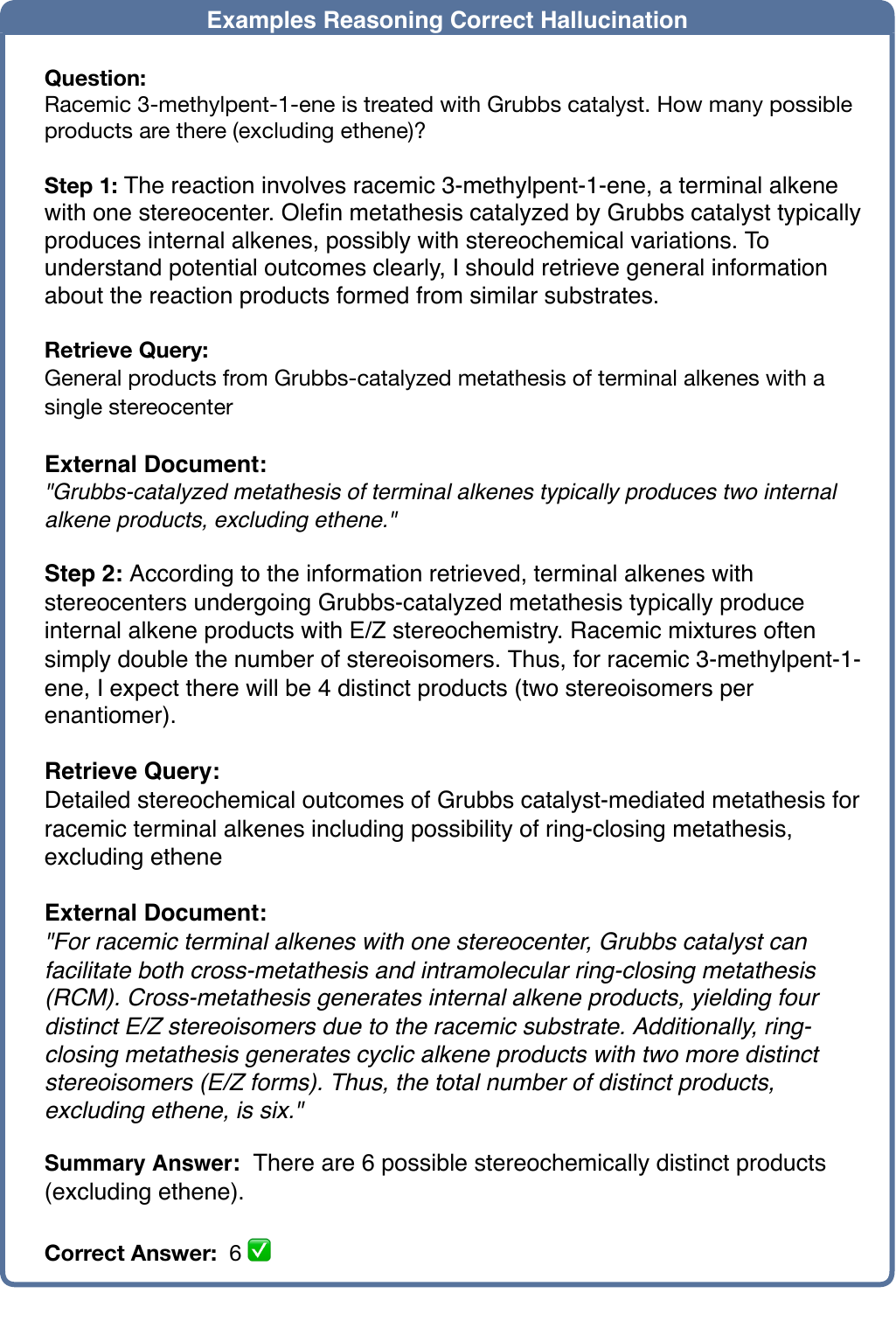} 
    \caption{
   Illustration of the effectiveness of MCTS-RAG. How further reasoning reduces retrieval-introduced hallucinations and improves accuracy.
    }
    \label{fig:correct_hallucination}
\end{figure}
\clearpage 
\begin{figure}[!h]
    \centering
    \begin{minipage}[b]{0.48\textwidth}
        \centering
        \includegraphics[width=\textwidth]{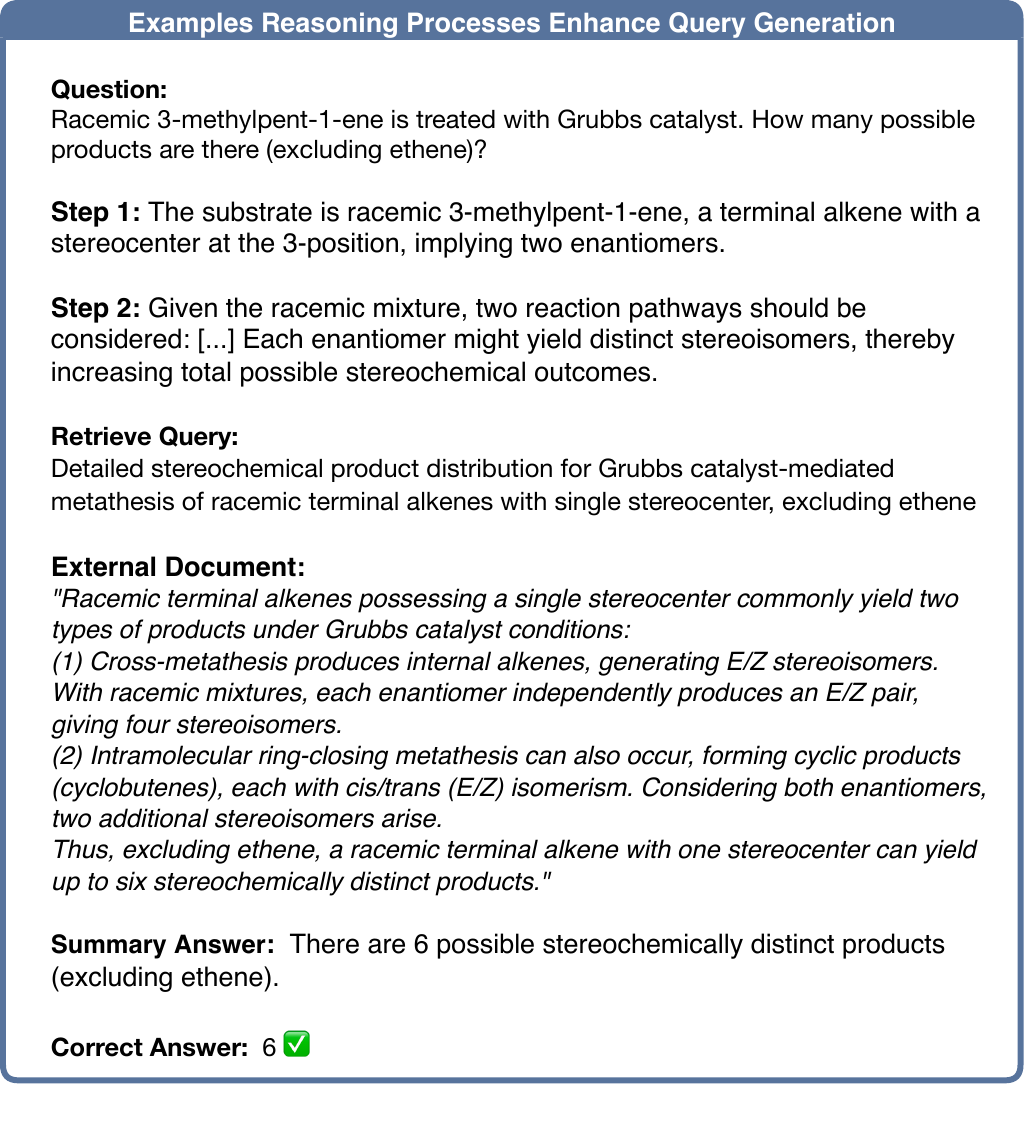}
        \caption{An illustration of the effectiveness of MCTS-RAG. Based on a clear chain of reasoning, it can generate higher quality retrieval queries and final answers, reduce hallucinations and improve accuracy.}
        \label{fig:reasoning_query}
    \end{minipage}\hfill 
    \begin{minipage}[b]{0.48\textwidth}
        \centering
        \includegraphics[width=\textwidth]{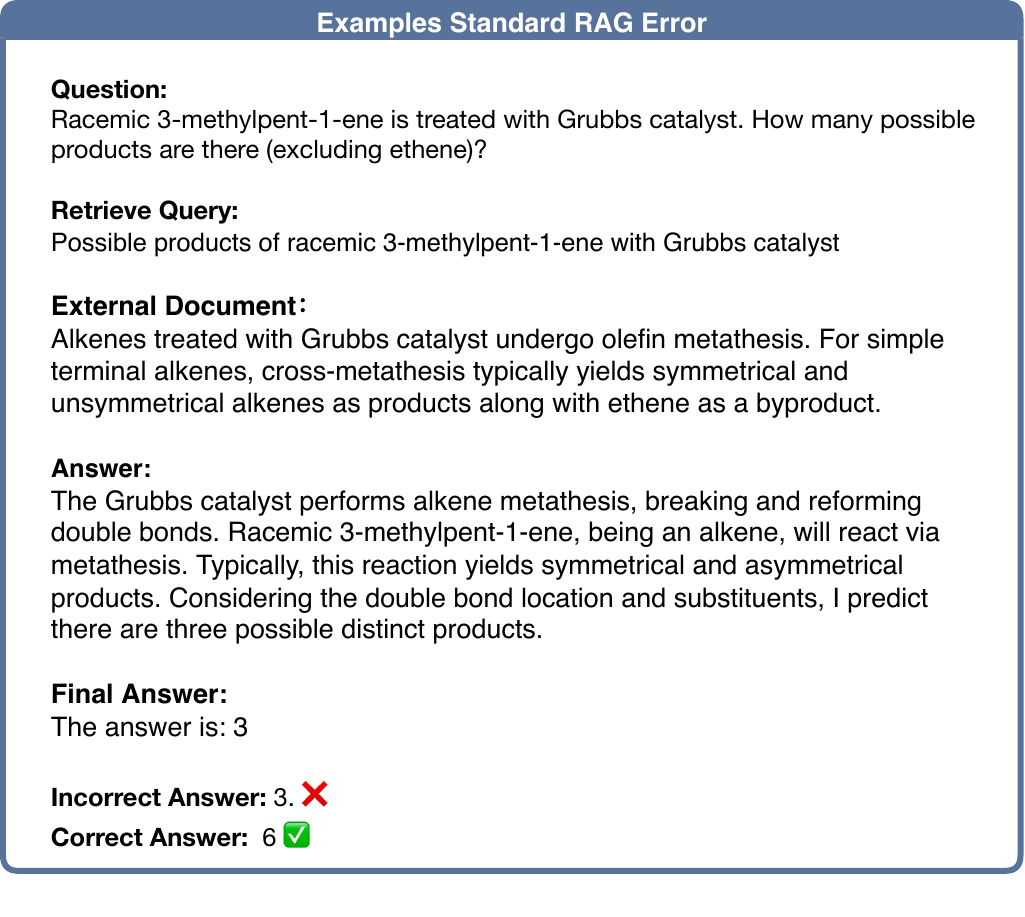}
        \caption{An illustration of standard RAG. Because the reasoning process is not clear enough, the final answer to the question is an illusion and the answer is wrong.}
        \label{fig:standard-rag-example}
    \end{minipage}
\end{figure}

\end{document}